%% file: Camera_Ready.tex
\documentclass[twoside]{article}

\usepackage[accepted]{aistats2021}
\usepackage[utf8]{inputenc} 
\usepackage[T1]{fontenc}    
\usepackage{url}            
\usepackage{booktabs}       
\usepackage{amsfonts}       
\usepackage{nicefrac}       
\usepackage{microtype}      

\usepackage{amsthm,amsmath} 

\usepackage[usenames,dvipsnames]{xcolor}
\usepackage[bookmarks=false]{hyperref}
\hypersetup{
  pdftex,
  pdffitwindow=true,
  pdfstartview={FitH},
  pdfnewwindow=true,
  colorlinks,
  linktocpage=true,
  linkcolor=Red,
  urlcolor=Red,
  citecolor=blue
}
\usepackage[capitalize]{cleveref} 
\usepackage{smile}
\usepackage{extarrows}
\usepackage[OT1]{fontenc}
\usepackage{comment}
\usepackage{stmaryrd}
\usepackage[dvipsnames]{xcolor}

\usepackage{algorithm}
\usepackage{algorithmic}

\RequirePackage{times}
\RequirePackage{natbib}
 
\definecolor{mypink1}{rgb}{0.858, 0.188, 0.478}
\definecolor{mypink2}{RGB}{219, 48, 122}
\definecolor{mypink3}{cmyk}{0, 0.7808, 0.4429, 0.1412}
\definecolor{mygray}{gray}{0.6}

\newcommand{\tmix}{t_{\text{mix}}}

\newcommand{\algname}{AAPI}

\usepackage{mathtools}

\usepackage[textsize=tiny]{todonotes}

\def\red#1{\textcolor{red}{#1}}

\begin{document}

%

%

\twocolumn[

\aistatstitle{Adaptive Approximate Policy Iteration}
\aistatsauthor{Botao Hao \And Nevena Lazic \And  Yasin Abbasi-Yadkori \And Pooria Joulani \And Csaba Szepesv\'ari }

\aistatsaddress{ Deepmind \And  Deepmind \And Deepmind \And Deepmind \And Deepmind} ]

\begin{abstract}
Model-free reinforcement learning algorithms combined with value function approximation have recently achieved impressive performance in a variety of application domains.
However, the theoretical understanding of such algorithms is limited, and existing results are largely focused on episodic or discounted Markov decision processes (MDPs). 
In this work, we present adaptive approximate policy iteration (AAPI), a learning scheme which enjoys a $\tilde{O}(T^{2/3})$ regret bound for 
undiscounted, continuing learning in uniformly ergodic MDPs. This is an improvement over the best existing bound of $\tilde{O}(T^{3/4})$ for the average-reward case with function approximation.  Our algorithm and analysis rely on online learning techniques, where value functions are treated as losses. The main technical novelty is the use of a data-dependent adaptive learning rate coupled with a so-called optimistic prediction of upcoming losses. In addition to theoretical guarantees, we demonstrate the advantages of our approach empirically on several environments.

\end{abstract}

\input{introduction}

\input{related}

\input{mdp}

\input{algorithm}

\input{analysis}

\input{experiments}

\input{discussion}

\subsubsection*{Acknowledgements}
Csaba Szepesv\'ari gratefully  acknowledges  funding  from 
the Canada CIFAR AI Chairs Program, Amii and NSERC.

\bibliographystyle{plainnat}
{\small
\bibliography{ref}
}

\setlength{\footskip}{55pt}

\clearpage
\onecolumn
\appendix
\begin{center}
    \Large Supplement to ``Adaptive Approximate Policy Iteration''
\end{center}

\input{supplement}

\end{document}

%% file: introduction.tex
\section{INTRODUCTION}

Our work focuses on model-free algorithms for learning in \emph{infinite-horizon undiscounted } Markov decision processes (MDPs), also known as average-reward MDPs. 
Although model-free algorithms have recently achieved impressive advances in multiple applications \citep{mnih2015human, van2016deep}, few performance guarantees exist, especially in the average-reward case with function approximation. 
In this work, we propose \emph{Adaptive Approximate Policy Iteration} (AAPI), a model-free learning scheme that can work with function approximation, and utilizes an adaptive data-dependent learning rate. 
We analyze the performance of AAPI in infinite-horizon undiscounted MDPs in terms of high-probability regret.

Our approach follows the ``online MDP'' line of work \citep{even2009online, neu2010online, lazic2019politex}, where the agent iteratively selects policies by running an 
online learning algorithm in each state, and the loss fed to each algorithm is the policy Q-function in that state. 
This results in a variant of approximate policy iteration (API), where the policy improvement step produces a policy optimal in hindsight w.r.t. \emph{the average of all previous} Q-functions rather than just the most recent one. 
The original work of \cite{even2009online} studied this scheme with known dynamics, tabular representation, and adversarial reward functions. More recent works \citep{lazic2019politex,mflq} have adapted this approach to the case of unknown dynamics, stochastic rewards, and value function approximation.  
The averaging of value functions is further justified theoretically and empirically by \cite{vieillard2019momentum} and \cite{vieillard2020leverage}.

A notable feature of our algorithm is that we exploit the fact that losses (Q-function estimates) are slow-changing.
In particular, our policy improvement step relies on the adaptive optimistic follow-the-regularized-leader (AO-FTRL) update \citep{mohri2016accelerating}.  
The resulting policies are Boltzmann distributions over the sum of past estimated Q-functions, coupled with an optimistic prediction of the upcoming loss and a state-dependent adaptive learning rate (softmax temperature). Our policy improvement step can also be seen as regularizing each policy by the KL-divergence to the previous policy; the reduction to online learning offers a principled way to scale such regularization.

On the theoretical side, we prove the first $\tilde{O}(T^{2/3})$ regret upper bound in the undiscounted, continuing setting with function approximation. This is an improvement over the best existing $\tilde{O}(T^{3/4})$ bound of
\cite{lazic2019politex} 
for the same setting,
which ignores the slow-changing nature of the estimated Q-functions. Our analysis exploits the fact that the change in consecutive Q-function estimates can be bounded by the change in policies. 
We rely on the results of \cite{rakhlin2013optimization}, but employ a different regret decomposition, with additional information provided by MDP properties. 
We emphasize that our learning framework is not limited to a particular function approximation method, and that in practice it serves the purpose of appropriately regularizing the policy improvement step of API.

%% file: related.tex
\textbf{Related work.} 
Most no-regret algorithms for infinite-horizon undiscounted MDPs are model-based, and only applicable to tabular representations \citep{bartlett09regal,jaksch2010near,ouyang2017learning,
fruit2018efficient,jian2019exploration,talebi2018variance}. 
In the model-free tabular setting, \cite{wei2019modelfree} show optimistic Q-learning achieves $O(\text{sp}(V_*)(XA)^{1/3}T^{2/3})$ regret in weakly-communicating MDPs, where $\text{sp}(V_*)$ is the span of the optimal state-value function, $X,A$ are the size of state and action spaces.
In the case of uniformly ergodic MDPs, they show a bound of $O(\sqrt{\tmix^3 \rho A T})$ on the \emph{expected regret}, where $\tmix$ is the mixing time and $\rho$ is the stationary distribution mismatch coefficient. In the model-free setting with function approximation, \cite{lazic2019politex} achieve $O(d^{1/2}T^{3/4})$ regret in ergodic MDPs.
Here $d$ is the size of the compressed state-action space ($XA$ for tabular representation, number of features for linear $Q$-functions). 

In episodic MDPs with horizon $H$, \cite{jin2018q} show an $O(\sqrt{H^3XAT})$ regret bound for Q-learning 
with tabular representation. With linear function approximation, \cite{yang2019reinforcement,jin2019provably, cai2019provably} show an $O(\sqrt{d^3H^3T})$ regret bound for an optimistic version of least-squares value/policy iteration under linear MDPs assumption.
The RLSVI algorithm \citep{osband16} performs exploration in the value function parameter space, and therefore can be applied with function approximation. Its worse-case regret bound of $O(\sqrt{H^5X^3AT})$ holds in the tabular setting \citep{russo2019worst} and $O(d^2\sqrt{H^4T})$ holds under the linear MDPs assumption \citep{zanette2020frequentist}. 

Another thread of the literature \citep{ross2011reduction, ross2014reinforcement} proposes a reduction of model-free RL to any no-regret online learning. While \cite{ross2011reduction} mainly focus on imitation learning, \cite{ross2014reinforcement} consider the finite-horizon case and uses a generic no-regret online learner that may result in a worse regret guarantee. Very recently, \cite{cheng2019predictor} exploit optimistic mirror descent to speed up policy optimization in RL, but do not provide a regret analysis in average-reward case. 

AAPI is also similar to the conservative policy iteration works, which attempt to stabilize API by regularizing each policy towards the previous policy \citep{kakade2002approximately,schulman2015trust,schulman2017proximal, abdolmaleki2018maximum,geist2019, vieillard2020leverage}. In particular, \cite{neu2017unified} identify several state-of-the-art entropy-regularized
RL algorithms as approximate variants of mirror descent, and \cite{shani2019adaptive} provides convergence rates for a mirror descent like algorithm in the discounted setting. \cite{vieillard2020leverage} provides a systematical analysis of regularization in RL.  To the best of the authors' knowledge, none of these works use adaptive data-dependent learning rate to accelerate policy learning.

%% file: mdp.tex
\section{PROBLEM SETTING}

We first introduce some notation. We use $\Delta_{\cS}$ to denote the space of probability distributions defined on the set $\cS$ and write $[d] = \{1,2,\ldots,d\}$. For vectors $u,v\in\mathbb R^d$, we define the weighted $\ell_2$-norm as $\|v\|_{u}^2=\sum_{i=1}^du_iv_i^2$ and $\ell_{\infty}$-norm as $\|u\|_{\infty}=\max_{j\in[d]}u_j$. In general, we treat discrete distributions as row vectors. 

 Infinite-horizon undiscounted MDPs are often characterized by a finite state space $\cX$,  a finite action space $\cA$, a reward function $r:\cX\times \cA\to[0, 1]$, and a transition probability function $P:\cX\times \cA \to \Delta_{\cX}$. The agent does not know the transition probability and the reward function in advance.  A policy $\pi:\cX\to\Delta_{\cA}$ is a mapping  from a state to a distribution over actions. Let $\{(x_t^{\pi}, a_t^{\pi})\}_{t=1}^{\infty}$
denote the state-action sequence obtained by following
policy $\pi$. The expected average reward of policy $\pi$ is defined as 
\begin{equation*}
   \lambda_{\pi}:=\lim_{T\to\infty}\mathbb E\left[\frac{1}{T}\sum_{t=1}^Tr(x_t^{\pi}, a_t^{\pi})\right]. 
\end{equation*}

The agent interacts with the environment as follows:  at each round $t$, the agent observes a state $x_t\in\cX$, chooses an action $a_t \sim \pi_t(\cdot | x_t)$, and receives
a reward $r(x_t,a_t)$. The environment then transitions to the next
state $x_{t+1}$ with probability $\mathbb P(x_{t+1}|x_t,a_t)$. The initial state $x_1$ is randomly generated from some unknown distribution.  Let $\pi^*$ be an unknown fixed policy. The regret of an algorithm with respect to this fixed policy is
defined as
\begin{equation}\label{def:regret}
    R_T= \sum_{t=1}^T \Big(\lambda_{\pi^*} - r(x_t, a_t)\Big)\,,
\end{equation}
where $a_t\sim \pi_t(\cdot|x_t)$.
The learning goal is to find an algorithm that minimizes the long-term regret $R_T$. Note that $R_T$ is still a random variable so we will bound it with high probability.

For each policy $\pi$, we denote $\cP^{\pi}\in\mathbb R^{|\cX|\times |\cX|}$ to be the Markov chain induced by $\pi$, where the component $(\cP^{\pi})_{x,x'}$ is the transition probability from $x$ to $x'$ under $\pi$, i.e. $(\cP^{\pi})_{x, x'} = \sum_{a\in \cA}\pi(a|x)P(x'|x,a)$. For a distribution $\mu$ over $\cX$, we let $\mu\cP^{\pi}$ be the distribution over $\cX$ that results from executing the policy $\pi$ for one step after the initial state is sampled from $\mu$. A stationary distribution $\mu_\pi$ of a policy $\pi$ over states satisfies $\mu_{\pi}\cP^{\pi}=\mu_{\pi}$. For a policy $\pi$, its expected reward can be expressed as 
\begin{equation*}
   \lambda_{\pi} = \mathbb E_{x\sim \mu_{\pi},a\sim \pi(\cdot|x)}\big[r(x, a)\big]. 
\end{equation*}

In this work, we focus on ergodic MDPs, a
sub-class of weakly communicating MDPs. 
An MDP is ergodic if the Markov chain induced by any policy $\pi$ is both irreducible and aperiodic, which means any state is reachable from any other state by following a suitable policy. 
It is well-known that all ergodic MDPs have an unique stationary state distribution, and so $\mu_\pi$ and $\lambda_\pi$ are well-defined. In addition, ergodic MDPs have a finite \emph{mixing time}, defined below.
\begin{definition}\label{def:mixing_time}
The mixing time of ergodic MDPs is defined as $\tmix:=$
\begin{equation*}
\max_{\pi}\min\left\{t\geq 1,  \Big\|(\cP^{\pi})^t(x, \cdot)-\mu_{\pi}\Big\|_1\leq \frac{1}{4}, \forall x\in\cX\right\},
\end{equation*}
that characterizes how fast MDPs reach stationary distributions from any
state under any policy.
\end{definition}

Finally, we define the value function under policy $\pi$ as
$$
V_{\pi}(x) = \mathbb E^{\pi}\Big[\sum_{t=1}^{\infty}(r(x_t, a_t) - \lambda_{\pi})|x_1 = x\Big]\,,
$$
where $\mathbb E^{\pi}$ is with respect to the sample path induced by $\pi$.
The state-action value function $Q_{\pi}(x,a)$ and $V_{\pi}(x)$ can also be defined as the unique solutions to the Bellman equation:
\begin{equation}\label{eqn:Bellman_eqn}
\begin{split}
     Q_{\pi}(x,a) &= r(x,a) -\lambda_{\pi} + \sum_{x'}P(x'|x, a)V_{\pi}(x')\\
     V_{\pi}(x) &= \sum_{a}\pi(a|x)Q_{\pi}(x, a).
\end{split}
\end{equation}

%% file: algorithm.tex
\section{Algorithm}\label{sec:alg}
\algname is a variant of approximate policy iteration and it proceeds in phases.  Suppose the total number of rounds is $T$. We divide $T$ into $K$ phases of length $\tau = T/K$ and assume $\tau$ is an integer for simplicity. Within each phase, our algorithm performs two tasks: policy evaluation and policy improvement. 

\textbf{Policy evaluation.}  In each phase $k\in[K]$, the algorithm executes the current policy $\pi_k$ for $\tau$ time steps, and computes an estimate $\hat{Q}_{\pi_k}$ of the true action-value function $Q_{\pi_k}$.
We leave unspecified the value function estimation method $\cG$; for example, one can use incremental algorithms, or both on-policy and off-policy data. AAPI is better interpreted as a learning schema. Our regret analysis will require that longer phase lengths lead to better estimates (made precise in Lemma~\ref{lemma:estimation_error}).  

\textbf{Policy improvement.} For each state $x\in\cX$, the policy improvement step takes the form of the adaptive optimistic follow-the-regularized-leader (AO-FTRL) update \citep{mohri2016accelerating}:
\begin{align}
    \label{eq:aoftrlmdp}
    \pi_{k+1}(a|x) = \argmax_{f\in\cF} 
    & \Big\langle f, \sum_{s=1}^{k} \hat{Q}_{\pi_s}(x,\cdot) + M_{k+1}(x, \cdot) \Big\rangle \notag \\
    & - \eta_k(x) \cR(f) \,.
\end{align}
(See Step 3 in Section~\ref{sec:proof_sketch} for a generic description of AO-FTRL.) The terms in Eq.~\eqref{eq:aoftrlmdp} are as follows:
\begin{itemize}
    \item The estimates $\hat Q_{\pi_s}(x,\cdot) \in\mathbb R^{|\cA|}$ are the loss functions fed to the AO-FTRL algorithm. $\cR(f)$ is the negative entropy regularizer, and $\cF$ is the probability simplex.
    \item The side-information $M_{k+1}(x, \cdot)\in\mathbb R^{|\cA|}$ is a vector computable based on past information and being predictive of the next loss $\hat Q_{\pi_{k+1}}(x,\cdot)$.  Since the policies are expected to change slowly due to the nature of exponential-weight-average type algorithms, we set $M_{k+1}(x,\cdot) = \hat{Q}_{\pi_k}(x,\cdot)$ (better guesses such as off-policy estimates can be used if available).
    \item The choice of learning rate $\eta_k(x)$ is crucial both theoretically and empirically. In particular, we choose $\eta_k(x)$ in a data-dependent fashion as
\begin{equation}\label{eqn:ada_learning}
    \eta_k(x) = \eta \sqrt{2\sum_{s=1}^k \|\hat{Q}_{\pi_s}(x,\cdot) - M_{s}(x, \cdot)\|_{\infty}^2} \,.
\end{equation}
A notable feature of $\eta_k(x)$ is that it is also state-dependent. Intuitively, for the choice $M_{s}(x,\cdot)= \hat{Q}_{\pi_{s-1}}(x,\cdot)$, the adaptive state-dependent learning rate results in a more exploratory policy for the states on which there is more disagreement between the past consecutive action-value functions.
\end{itemize}
Based on \eqref{eq:aoftrlmdp}, the next policy is a Boltzmann distribution (a consequence of negative entropy regularizer) over the sum of all past state-action value estimates and the side-information:
\begin{equation}\label{eqn:main_update}
    \pi_{k+1}(a|x)\propto \exp\Big(\eta_k^{-1}(x)\big(\sum_{s=1}^{k} \hat{Q}_{\pi_s}(x,a) + M_{k+1}(x, a)\big)\Big).
\end{equation}
\begin{remark}
\algname is similar to the \textsc{Politex} algorithm \cite{lazic2019politex}, where the main difference is that \textsc{Politex} sets the next policy to $\pi_{k+1}(a|x) \propto \exp(\eta^{-1} \sum_{s=1}^k \widehat Q_{\pi_s})$ in the improvement step. We demonstrate that the use of side-information and adaptive learning rates improves both the theoretical guarantees (Theorem \ref{thm:main}) and empirical performance (Section \ref{sec:exp}) over \textsc{Politex}. The overall algorithm is summarized in Algorithm \ref{alg:ada_politex}.
	\vspace{-0.1in}
\end{remark}

\begin{algorithm}[htb!]
\caption{Adaptive approximate policy iteration (AAPI)}
\begin{algorithmic}[1]\label{alg:ada_politex}
		\STATE
		\textbf{Input:} phase length $\tau$, number of phase $K$,  initial state $x_0$, turning parameter $\eta$, value function estimation algorithm $\cG$.
		\STATE
		\textbf{Initialize:} $\pi_1(a|x) = 1 / |\mathcal{A}|$, $ \forall x, a$;
		\STATE
		\textbf{Repeat:}
		\FOR{$k=1,\ldots, K$}
			\STATE Execute $\pi_k$ for $\tau$ time steps and collect dataset $\cD_k$. 
			\STATE Estimate $\hat{Q}_{\pi_k}$ from $\cD_1, \ldots, \cD_k$ using $\cG$.
			\STATE Calculate adaptive learning rate:
			\begin{equation*}
  \eta_k(x) = \eta \sqrt{2\sum_{s=1}^k \|\hat{Q}_{\pi_s}(x,\cdot) - M_{s}(x, \cdot)\|_{\infty}^2},
  \end{equation*}
   where   $M_{s} = \hat{Q}_{\pi_{s-1}}$.
   \STATE Calculate 
   \begin{equation*}
       \text{index}(x,a) = \sum_{s=1}^{k} \hat{Q}_{\pi_s}(x,a) + M_{k+1}(x, a).
   \end{equation*}
  \STATE Update next policy as:
  \begin{equation*}
    \pi_{k+1}(a|x) \propto \exp\Big(\eta_k(x)^{-1} \text{index}(x,a)\Big),
\end{equation*}
		where $\eta_k(x)$ is defined in Eq.~\eqref{eqn:ada_learning}.
		\ENDFOR
		
\STATE		\textbf{Output:} $\pi_{K+1}$
	\end{algorithmic}
\end{algorithm}

%% file: analysis.tex
\section{ANALYSIS}

To derive a regret bound for Algorithm~\ref{alg:ada_politex}, we decompose the cumulative regret \eqref{def:regret} as follows:
\begin{equation}\label{eqn:regret_dec}
    \begin{split}
        R_T &= 
 \sum_{t=1}^T\Big(\lambda_{\pi_t}- r(x_t, a_t)\Big) + \sum_{t=1}^T\Big(\lambda_{\pi^*}-\lambda_{\pi_t}\Big).
    \end{split}
\end{equation}
The first term captures the sum of differences between observed rewards and their long term averages. If policies are changing
slowly, or if they are kept fixed for extended periods of time,
we expect this term to capture the noise in the regret. The second term is called \emph{pseudo-regret} in literature. It measures the difference between the expected reward of a fixed policy and the policies produced by the algorithm.

We first impose a condition on the quality of policy evaluation at each phase. For a probability distribution $\mu$ on $\cX$ and a stochastic policy $\pi$, define $\mu\otimes \pi$ to be the distribution on $\cX\times \cA$ that puts the probability mass $\mu(x)\pi(a|x)$ on pair $(x,a)\in\cX\times \cA$. Recall that $\mu_{\pi^*}$ is the stationary distribution of $\pi^*$ over the states.
\begin{condition}\label{con:estimation_error}
For each phase $k\in[K]$, denote $D_{\pi_k} = \hat{Q}_{\pi_k}-Q_{\pi_k}$. We assume the following holds with probability $1-\delta$,
\begin{equation}
    \begin{split}
        &\max\Big\{\|D_{\pi_k}\|_{\mu_{\pi^*}\otimes \pi^*}, \|D_{\pi_k}\|_{\mu_{\pi^*}\otimes \pi_k}, \|D_{\pi_k}\|_{\infty}\Big\}\\
        &\leq \varepsilon_0+\tilde{C}\sqrt{\frac{\log(1/\delta)}{\tau}},
    \end{split}
\end{equation}
where $\varepsilon_0$ is the irreducible approximation error and $\tilde{C}$ is a problem dependent constant.  Additionally, there exists a constant $b$ such that $ \hat{Q}_{\pi_k}(x,a)\in[b,b+Q_{\max}]$ for any pair $(x,a)\in\cX\times \cA$ and $k\in[K]$.
\end{condition}

\begin{remark}
The problem dependent constant $\tilde{C}$ will in general depend on $d, \tmix, \mu_{\pi^*}, \mu_{\pi_k}$. Here, $d$ is the dimension of the representation (e.g. $|\cX||\cA|$ for the tabular case, or number of features for the linear value function case). 
\end{remark}

\begin{remark}
The requirement for the $\mu_{\pi^*}\otimes \pi^*$-norm and $\mu_{\pi^*}\otimes \pi_k$-norm has been shown to hold, for example, with linear value function approximation using the LSPE algorithm \citep{bertsekas1996temporal}, under Assumptions \ref{assum:linear_independent}-\ref{assumption:weighted-error} given in the Appendix. 
Lemma~\ref{lemma:1} in the Appendix shows that the requirement for $\ell_{\infty}$-norm can also be satisfied, for example, with linear value functions, under similar conditions.
\end{remark}
\begin{remark}
The estimation error generally depends on the mismatch between distributions $\mu_{\pi_k}$ and $\mu_{\pi^*}$.  With value functions linear in features $\phi(x, a) \in \mathbb R^d$, this mismatch depends on the spectra of matrices $ \mathbb E_{\nu}[\phi(x, a) \phi(x, a)^\top]$ for different distributions $\nu$, and need not scale in the number of state-action pairs. See Assumption A4 in \cite{lazic2019politex} for a more detailed explanation.
\end{remark}

\begin{theorem}[Main result]\label{thm:main}
Consider an ergodic MDP and suppose Condition \ref{con:estimation_error} holds. By choosing the phase length 
$\tau = (\tilde{C}/\rho\tmix^3)^{2/3}T^{2/3}$, we have with probability at least $1-1/T$,
\begin{equation*}
    R_T = \tilde{O}\left(\tmix^2(\rho\tilde{C}^2)^{1/3}T^{2/3} + T\varepsilon_0\right),
\end{equation*}
where $\rho$ is the distribution mismatch coefficient that has used in previous
work \citep{kakade2002approximately, agarwal2020optimality, wei2019modelfree} and $\tilde{\cO}(\cdot)$ hides universal constants and poly-logarithmic factors. 
\end{theorem}

\begin{remark}
It is worth comparing the above result with the regret bound presented in \cite{lazic2019politex}. Ignoring the irreducible error $\varepsilon_0$, we improve the leading order of their general result (Corollary 4.6 in \cite{lazic2019politex}) from $\tilde{O}(T^{3/4})$ to $\tilde{O}(T^{2/3})$. When  specialized to linear value function approximation where $\tilde{C}$ scales with $d^{1/2}$ (Theorem 5 in \cite{lazic2019politex}), we improve their results from $\tilde{O}(d^{1/2}T^{3/4})$ to $\tilde{O}(d^{1/3}T^{2/3})$. 
\end{remark}
\begin{remark}
It is worth to mention that \cite{wei2019modelfree} obtains  $\tilde{O}(\sqrt{T})$ regret in terms of \emph{expected regret} in the tabular case for ergodic MDPs while we consider \emph{high-probability regret}. In particular, their analysis does not account for the estimation and approximation errors in Q-functions that will significantly complicate the analysis and result in a worse regret bound.
\end{remark}

\section{PROOF SKETCH}
\label{sec:proof_sketch}

In this section, we provide a proof sketch for Theorem \ref{thm:main}. Technical details are deferred to Appendix~\ref{sec:proof_main}. At a high level, we bound the two terms in the regret decomposition Eq.~\eqref{eqn:regret_dec} separately. While the first term is bounded by the fast mixing condition, the second term is split into the regret due to value function estimation error and the regret due to online learning reduction.

\textbf{Step 1: fast mixing.} To bound the first term in Eq.~\eqref{eqn:regret_dec}, we require the following uniform fast mixing condition, which is used frequently in online MDP literature \citep{even2009online,neu2010online}. Note that ergodic MDPs that this paper focuses on automatically satisfy this condition.
\begin{condition}[(Uniform fast mixing)]\label{con:mixing}
There exists a number $\tmix>0$ such that for any policy $\pi$ and any pair of distributions $\mu$ and $\mu'$ over $\cX$, it holds that
\begin{equation}
    \big\|(\mu-\mu')\cP^{\pi}\big\|_1\leq \exp(-1/\tmix)\big\|\mu-\mu'\big\|_1.
\end{equation}
\end{condition}

The following lemma provides upper bounds for the first term (see e.g. Lemma 4.4 in \cite{lazic2019politex} for a proof). 
\begin{lemma} \label{lemma:bound_VW}
Suppose that Condition \ref{con:mixing} holds. The following inequality holds with probability at least $1-\delta$,
\begin{equation*}
    \begin{split}
        &\Big|\sum_{t=1}^T\big(\lambda_{\pi_t} - r(x_t, a_t)\big)\Big|\\
 &\leq K\tmix + 4\sqrt{2}\tmix\sqrt{KT\log (T/\delta)}, 
    \end{split}
\end{equation*}
where $K$ is the number of phases.
	\vspace{-0.1in}
\end{lemma}

\textbf{Step 2: decomposition.} We bound the second term (pseudo regret) in Eq.~\eqref{eqn:regret_dec}. Since the policy is only updated at the end of each phase of length $\tau$ (see line 9 in Algorithm \ref{alg:ada_politex}), we have $\pi_t = \pi_k$ for $t\in\{\tau (k-1), \ldots, \tau k\}$. Thus, the pseudo-regret term can be rewritten as
\begin{eqnarray}\label{eqn:per_diff}
\sum_{t=1}^T\Big(\lambda_{\pi^*}-\lambda_{\pi_t}\Big) = \tau\sum_{k=1}^K\Big(\lambda_{\pi^*}-\lambda_{\pi_k}\Big).
\end{eqnarray}

We slightly abuse the notation by writing $Q_{\pi}(x, \pi') = \sum_{a}\pi'(a|x)Q_{\pi}(x, a)$. In particular, $Q_{\pi}(x, \pi)$ is exactly the value function $V_\pi(x)$ by Definition \ref{eqn:Bellman_eqn}.  Applying the performance difference lemma (Lemma \ref{lemma_per_diff} in the supplementary material), we have
\begin{align*}
    \lambda_{\pi^*} -\lambda_{\pi_k} &= \Big\langle \mu_{\pi^*}, Q_{\pi_{k}}(\cdot, \pi_{*})-Q_{\pi_{k}}(\cdot, \pi_k) \Big\rangle.
\end{align*}
Bridging by empirical estimations, we decompose \eqref{eqn:per_diff} into $R_{1T} + R_{2T}$, where
\begin{equation}\label{eqn:decom_I1}
    \begin{split}
        R_{1T}&= \tau\sum_{k=1}^K \Big\langle \mu_{\pi^*}, Q_{\pi_{k}}(\cdot, \pi_{*})-\hat{Q}_{\pi_{k}}(\cdot, \pi_{*}) \Big\rangle\\ 
        &+ \tau\sum_{k=1}^K\Big\langle \mu_{\pi^*}, \hat{Q}_{\pi_{k}}(\cdot, \pi_k)-Q_{\pi_{k}}(\cdot, \pi_k) \Big\rangle,\\
 R_{2T}&=
\tau\sum_{k=1}^K\Big\langle \mu_{\pi^*}, \hat{Q}_{\pi_{k}}(\cdot, \pi^*)-\hat{Q}_{\pi_{k}}(\cdot, \pi_k)\Big\rangle. 
    \end{split}
    	\vspace{-0.1in}
\end{equation}

\textbf{Step 3: estimation error.} The term $R_{1T}$ quantifies the regret incurred in the policy evaluation step due to the estimation error and function approximation error of Q-function in each phase. It can be bounded as in Theorem 4.1 of \cite{lazic2019politex} under similar assumptions, which we reproduce here for completeness.  
\begin{lemma}\label{lemma:estimation_error}
Suppose Condition \ref{con:estimation_error} holds. Then 
\begin{equation}\label{bound:R_1T}
    R_{1T}\leq T\Big(\varepsilon_0 + \tilde{C}\sqrt{\frac{\log(1/\delta)}{\tau}}\Big),
\end{equation}
with probability at least $1-\delta$. 
	\vspace{-0.1in}
\end{lemma}


\textbf{Step 4: online learning reduction.} Minimizing $R_{2T}$ can be cast into an online learning problem \citep{cesa2006prediction, shalev2012online}, and this observation determines the choice of our algorithm. Previous work has tackled this subproblem using mirror descent, resulting in $\tilde{O}(T^{3/4})$ regret after optimizing $\tau$ ignoring the irreducible error $\varepsilon_0$. Here we instead use the AO-FTRL framework, which allows us to show an improved $\tilde{O}(T^{2/3})$ regret bound. As we show, the reason we can benefit from optimism is that the losses (Q-functions) change slowly, and we carefully transfer this knowledge to the adaptive learning rate. This is the main technical contribution of the paper.

 First, we state the framework of AO-FTRL and its regret results. Let $\{q_t\}_{t=1}^T$ be a sequence of loss vectors and let $\{f_t\}_{t=1}^T\subseteq{\cF}$ be a sequence of prediction vectors, where $\cF$ is the probability simplex. At the beginning of each round, the algorithm receives a side-information vector $M_t$. In literature, $\{M_s\}_{s=1}^t$ are also called predictable sequences \citep{rakhlin2012online}, and the algorithm can be seen as a way of
utilizing prior knowledge about loss sequences.  The algorithm then selects an action $f_t$, and suffers a cost $\langle f_t, q_t \rangle$.  The goal of this online learning problem is to minimize the cumulative regret with respect to the best action in hindsight $f^*$, defined as $\tilde{R}_T = \sum_{t=1}^T \langle f_t - f^*, q_t \rangle$\,.

Let $\cR:\cF\to\mathbb R$ be a 1-strongly convex regularizer on $\cF$ with respect to some norm $\|\cdot\|$ and denote by $\|\cdot \|_{*}$ its dual norm. Initialize $f_1=\argmin_{f\in\cF}\cR(f)$. At each round $t$,
 AO-FTRL has the following form:
\begin{equation*}
\begin{split}
     &f_{t+1} = \argmin_{f\in\cF}\langle f, \sum_{s=1}^t q_s + M_{t+1}\rangle + \eta_t \cR(f),\\
     &\eta_t =\eta\sqrt{\sum_{s=1}^t\|q_s-M_s\|^2_{*}}, 
\end{split}
\end{equation*}
where $\eta$ is an absolute constant. It's easy to see that $\eta_t$ is non-decreasing. For simplicity, we assume $M_1 = 0, \eta_0 = 0$. Next lemma provides a generic regret bound for AO-FTRL. The detailed proof is deferred to Appendix \ref{proof:FTRL} in the supplementary material.

\begin{lemma} \label{lemma:FTRL}
Choose $\eta = \sqrt{2/\cR(f^*)}$ and denote $R_{\max}=\max_f\cR(f)$. The cumulative regret for AO-FTRL is upper-bounded by
\begin{equation}\label{eqn:online_regret}
\begin{split}
     &\tilde{R}_T\leq \sqrt{2R_{\max}\sum_{t=1}^T\|q_t-M_t\|^2_{*}}\\
     &-\sum_{t=1}^{T}\frac{\eta_{t}}{4}\|f_t-f_{t+1}\|^2 + \langle M_{T+1}, f^*-f_{T+1}\rangle.
\end{split}
\end{equation}
\end{lemma}

\begin{remark}
Unlike the AO-FTRL analyses of \cite{rakhlin2012online, mohri2016accelerating},
but similarly to, e.g., the analysis of \cite{joulani2017modular}, Eq.~\eqref{eqn:online_regret} has a key negative term (at the expense of a slightly larger constant factor in the main positive term). These negative terms, which are retained from a tight regret bound on the forward regret of AO-FTRL \citep{joulani2017modular}, track the evolution of the policy $f_t$. With the proper choice of $M_t$, the norm terms $\| q_t - M_t \|_{*}$ will also be controlled by the evolution of $f_t$ (see Lemma \ref{lemma:Q_error}), and the aforementioned negative terms allow us to greatly reduce the contribution of the norm terms $\| q_t - M_t \|_{*}$ to the overall regret.
\end{remark}

The reason that minimizing $R_{2T}$ can be cast into an online learning problem is as follows. By the definition of $Q_{\pi}(x, \pi')$ in Step 2, we rewrite $R_{2T}$ in \eqref{eqn:decom_I1} as
\begin{equation*}
    \begin{split}
        R_{2T} = \tau \sum_{x\in\cX}\mu_{\pi^*}(x)\sum_{k=1}^K  \Big\langle \pi^*(\cdot|x)-\pi_k(\cdot|x), \hat{Q}_{\pi_{k}}(x, \cdot)\Big\rangle.
    \end{split}
\end{equation*}
For each state $x\in\cX$, we view $\pi_k(\cdot|x)$ as the prediction vector and $\hat{Q}_{\pi_k}(x,\cdot)$ as the loss vector. The equivalence between $R_{2T}$ and $\tilde{R}_T$ enables us to utilize the generic regret bound for AO-FTRL in Lemma \ref{lemma:FTRL} for each individual state.
%
%

Next, we will show that under some conditions, the change in the true Q values can be bounded by the change of policies. This is a unique property of ergodic MDPs that allows us to benefit from the negative term in \eqref{eqn:online_regret}. To ensure $Q_{\pi}$ is unique, we assume $\sum_x \mu_{\pi}(x)V_{\pi}(x) = 0$.
\begin{lemma}[Relative Q-function Error]\label{lemma:Q_error}
For any two successive policies $\pi_{k-1}$ and $\pi_k$, the following holds for any state-action pair $(x, a)$,
\begin{equation*}
\begin{split}
    &\Big|Q_{\pi_k}(x,a) - Q_{\pi_{k-1}}(x, a)\Big|\\
   & \leq \tmix^2\log_2^2(K) \max_x\big\|\pi_{k-1}(\cdot|x)-\pi_k(\cdot|x)\big\|_1 + \frac{2}{K^3}.
\end{split}
\end{equation*}

\end{lemma}
The detailed proof of Lemma \ref{lemma:Q_error} is deferred to Appendix~\ref{proof:relative_Q}. Combining the result in Lemmas \ref{lemma:FTRL} and \ref{lemma:Q_error}, we can derive the following lemma. 

\begin{lemma}\label{thm:R_2T}
 Suppose Condition \ref{con:estimation_error} holds. Then the following upper bound holds with probability at least $1-\delta$,
\begin{equation}\label{eqn:online_regret1}
\begin{split}
     R_{2T}
     \lesssim\tau\tmix^4\rho\log_2^4(K) + T\Big(\frac{\tilde{C}^2\log (1/\delta)}{\tau} + \varepsilon_0^2\Big),
\end{split}
\end{equation}
where $\lesssim$ hides universal constant factors.
\end{lemma}
The detailed proof of Lemma \ref{thm:R_2T} is deferred to Appendix~\ref{proof:thm:R_2T} in the supplementary material. Finally, we optimize $\tau$ to be $(\tilde{C}/\rho\tmix^3)^{2/3}T^{2/3}$ and reach our conclusion.

\begin{remark}
Within the upper bound \eqref{eqn:online_regret1}, $\tilde{C}^2\log (1/\delta)/\tau + \varepsilon_0^2$ stands for the approximation error and estimation error per round. When value functions can be computed exactly (known MDP) and for phase length $\tau=1$, the online learning reduction regret for AAPI scales logarithmically in the number of phases $K$, while {\textsc POLITEX} \citep{lazic2019politex} scales as $\sqrt{K}$. This is the main reason that we can improve the regret from $\tilde{O}(T^{3/4})$ to $\tilde{O}(T^{2/3})$. 
	\vspace{-0.1in}
\end{remark}

%% file: experiments.tex
\section{EXPERIMENTS}\label{sec:exp}

In this section we provide an empirical evaluation of \algname on several environments. We compare \algname to POLITEX, which corresponds to updating policies using a mirror descent rule rather than AO-FTRL. 
We also evaluate RLSVI (\citep{osband16}, Algorithms 1 and 2 
with $\sigma^2=1$ and tuned $\lambda$), where policies are greedy w.r.t. a randomized estimate of $Q_{*}$.  
Overall, we find that AAPI performs well in discrete-state environments such as DeepSea \citep{osband2017deep}, whereas the adaptive per-state learning rate is less helpful in environments such as CartPole \citep{barto1983neuronlike} with continuous states and smooth dynamics.

\input{experiments_supp}

%% file: experiments_supp.tex
\begin{figure*}[!t]
 \centering
\includegraphics[width=0.8\linewidth, trim=1.5cm 0.1cm 2.8cm 0cm, clip]{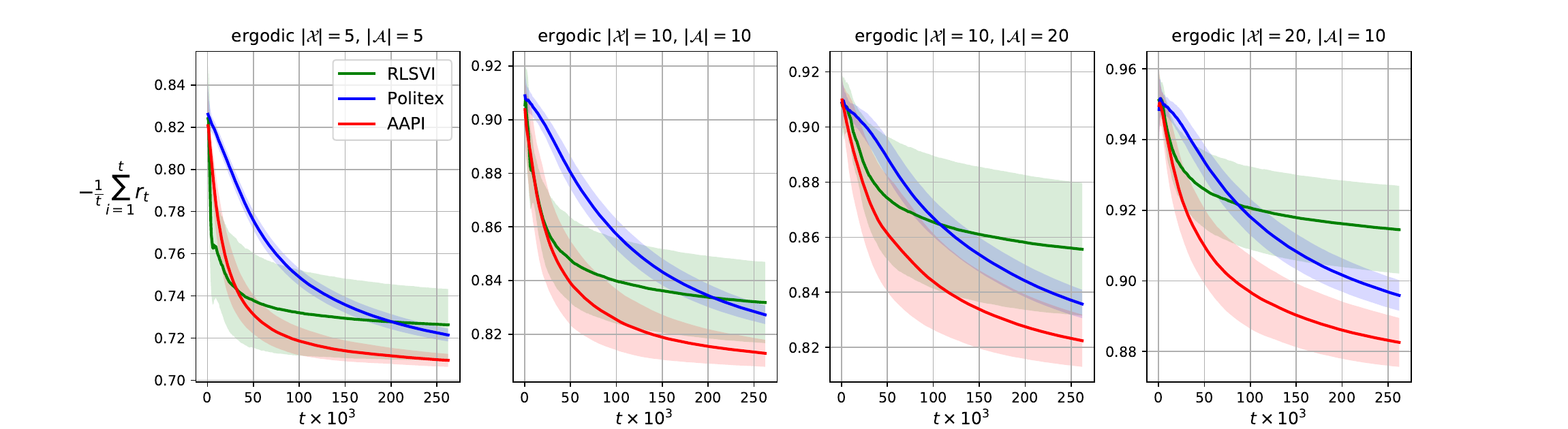}
\caption{Evaluation on a tabular ergodic MDP.}
\label{fig:ergodic}
\end{figure*}

 \begin{figure*}[!t]
 \centering
\includegraphics[width=0.8\linewidth, trim=1.5cm 0.1cm 2.8cm 0cm, clip]{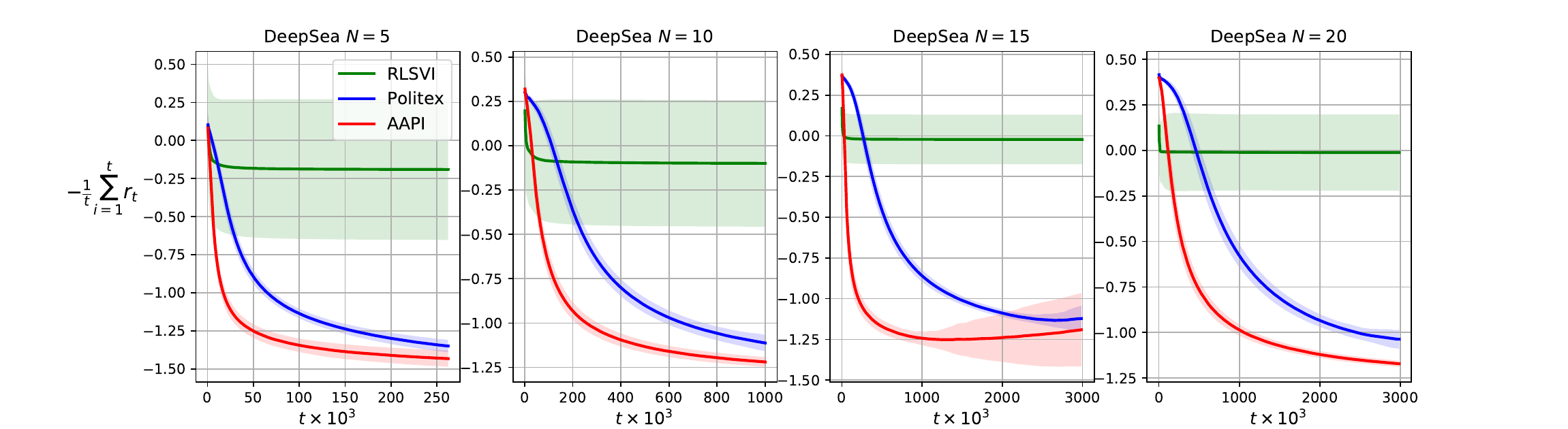}
\caption{Evaluation on DeepSea environments of different sizes.}
\label{fig:deepsea}
\end{figure*}

 \begin{figure}[!t]
 \centering
\includegraphics[width=0.95\linewidth, trim=0.1cm 0cm 2.2cm 0cm, clip]{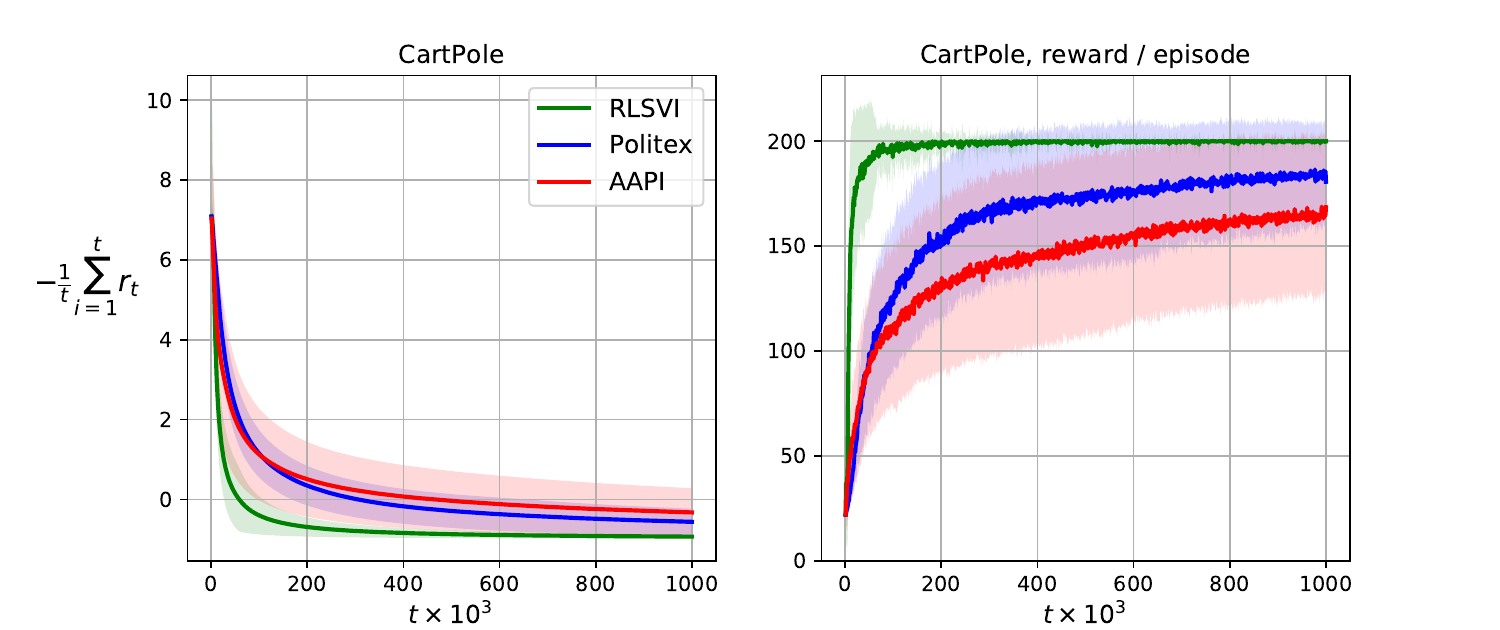}
\caption{Evaluation on the CartPole environment.}
\label{fig:cartpole}
\end{figure}
We approximate all value functions using least-squares Monte Carlo, i.e. linear regression from state-action features to empirical returns. For MDPs with a large or continuous state space $\cX$, updating per-state learning rates can be impractical. 
 Instead, we store the weights of past Q-functions in memory, and for each state in the trajectory, we compute the learning rate using a subset of $n_k \leq 30$ randomly-selected past weight vectors (we correct the scale of the estimate by multiplying with $\sqrt{k / n_k}$. With rich function approximation such that neural networks, one can keep a fixed buffer with a subset of the previous Q-functions (chosen in a randomized way, or keeping the most recent K networks as in \cite{lazic2019politex}), or train distillation networks that summarize the sum of previous Q-functions. Another possibility is to parameterize $\pi_k$ and optimize the objective w.r.t. the parameters. For Boltzmann policies, we tune the constant $\eta$ for the learning rate $\eta_k(x)$ in the range $[0.01, 100]$.
For each environment and algorithm we evaluate $-\sum_{s=1}^t r_t/t$ and plot the mean and standard deviation over 50 runs.  The environments we evaluate on are as follows. 

 \textbf{Tabular ergodic MDPs.} We consider a simple tabular MDP where $r(1, a)=1$, $r(x, a)=0$ for $x \neq 1$. On any action in state 1, the environment transitions to a randomly chosen state $x \neq 1$. On action 1 in a state $x \neq 1$, the environment transitions to state $x-1$ with probability 0.9, and to a randomly chosen state with probability 0.1. On all other actions in $x \neq 1$, the environment transitions to a randomly chosen state. We represent state-action pairs using one-hot indicator vectors of size $|\cX|| \cA|$, and experiment with different sizes of the state and action spaces $\cX$ and $\cA$.

\textbf{DeepSea} \citep{osband2017deep}.
In the DeepSea environment, states comprise an $N \times N$ grid, and there are two actions. The environment transitions and costs are deterministic. 
The agent starts in the top-left cell $(0, 0)$. 
On action 0, the agent transitions down and left, and receives reward 0. On action 1, the agent transitions down and right, and receives reward -1. On transitioning to the bottom-right cell $(N-1, N-1)$, the agent receives reward $2N$.  The infinite-horizon version of the environment wraps the environment around the vertical axis. 
An optimal strategy first takes the action 1 $N$ times (to get to $(N-1, N-1)$) and then takes an equal number of 0 and 1 actions, and has expected average reward close to $1.5$.  A simple strategy that always takes action 1 has an average reward $1$, and a suboptimal strategy that only takes action 0 has an average reward of $0$. 
We represent states as length-$2N$ vectors containing one-hot indicators for each grid coordinate, and estimate linear $Q$-functions.

\textbf{CartPole} \citep{barto1983neuronlike}.
In the CartPole environment, the goal is to balance an inverted pole attached by an unactuated joint to a cart, which moves along a frictionless rail.  There are two actions, corresponding to pushing the cart to the left or right. The observation consists of the position and velocity of the cart, pole angle, and pole velocity at the tip. 
There is a reward of +1 for every timestep that the pole remains upright. The episodic version of the environment ends if the pole angle is more than 15 degrees from vertical, if the cart moves more than 2.4 units from the center, or after 200 steps. In the infinite-horizon version, if the episode ends after $h$ steps, we return a reward of $h - 200$ and reset. 
For this task, in addition to the given observation, we extract multivariate Fourier basis features \cite{konidaris2011value} of order 4.

\textbf{Discussion.}  In most of our experiments, adaptive learning rate speeds up the convergence of approximate policy iteration, compared to using a constant learning rate as in Politex.
The adaptive per-state learning rate is less helpful in CartPole, possibly because observations are continuous and dynamics are smooth, so there is higher generalization across states.

%% file: discussion.tex
\section{CONCLUSION}

We have presented \algname, a model-free learning scheme that can work with function approximation, and enjoys a $\tilde{O}(T^{2/3})$ regret guarantee in infinite-horizon undiscounted, ergodic MDPs. \algname improves upon previous results for this setting by using the slow-changing property of policies in both theory and practice. 
One direction for future work is improving the policy evaluation stage. While we estimate each value function solely using the $\tau$ on-policy transitions, better estimates can potentially be obtained using all data. Using more sophisticated side-information, such as a weighted average of past Q-estimates or an off-policy estimate of the Q-function may also be helpful in practice.   Other future work may include practical implementations of the algorithm when trained with neural networks that maintain only a subset of past networks in memory; one possible practical approach is given by \citet{vieillard2020leverage}.

%% file: supplement.tex
\input{ftrl.tex}

\subsection{Proof of Lemma \ref{thm:R_2T}: online learning reduction} \label{proof:thm:R_2T}

\textbf{Step 1.} We utilize Lemma \ref{lemma:FTRL} for each individual state $x$. Recall that 
\begin{eqnarray*}
  R_{2T} &=& \tau\sum_{k=1}^K\Big\langle \mu_{\pi^*}, \hat{Q}_{\pi_{k}}(\cdot, \pi^*)-\hat{Q}_{\pi_{k}}(\cdot, \pi_k)\Big\rangle\\
  &=& \tau \sum_{x\in\cX}\mu_{\pi^*}(x)\sum_{k=1}^K  \Big\langle \pi^*(\cdot|x)-\pi_k(\cdot|x), \hat{Q}_{\pi_{k}}(x, \cdot)\Big\rangle.
\end{eqnarray*}
Applying Lemma \ref{lemma:FTRL} with $f_k = \pi_k(\cdot|x)$, $q_k = \hat{Q}_{\pi_k}(x,\cdot)$ and $M_k = \hat{Q}_{\pi_{k-1}}(x,\cdot)$, we have
\begin{equation}\label{eqn:R1_bound_conti}
\begin{split}
        R_{2T}\leq \tau\sum_{x\in\cX}\mu_{\pi^*}(x)&\Big(\sqrt{2R_{\max}} \sqrt{\sum_{k=1}^K\big\|\hat{Q}_{\pi_k}(x, \cdot)-\hat{Q}_{\pi_{k-1}}(x, \cdot)\big\|_{\infty}^2}\\
    &-\sum_{k=1}^K\frac{\eta_k(x)}{4}\big\|\pi_k(\cdot|x)-\pi_{k+1}(\cdot|x)\big\|_{1}^2 + 2(b+Q_{\max})\Big),
    \end{split}
\end{equation}
since $\hat{Q}_{\pi_K}(x,a)\in[b,b+Q_{\max}]$ from Condition \ref{con:estimation_error}. Here, $\eta_k(x) =\eta \sqrt{\sum_{s=1}^k \|\hat{Q}_{\pi_s}(x,\cdot) - \hat{Q}_{\pi_{s-1}}(x,\cdot)\|_{\infty}^2}$. 

\textbf{Step 2.} It remains to bound the cumulative change of estimated $Q$-values in Eq.~\eqref{eqn:R1_bound_conti}. We first decompose it by substrating the true $Q$-function and using the triangle inequality and $2ab \leq a^2 + b^2$:
\begin{equation}\label{eqn:Q_diff_dec}
    \begin{split}
       \sum_{k=1}^K \big\|\hat{Q}_{\pi_k}(x, \cdot) - \hat{Q}_{\pi_{k-1}}(x, \cdot)\big\|_{\infty}^2
\leq&\sum_{k=1}^K 2\big\|\hat{Q}_{\pi_k}(x, \cdot)- Q_{\pi_k}(x, \cdot)\big\|_{\infty}^2+\sum_{k=1}^K 2\big\|Q_{\pi_{k-1}}(x, \cdot)- \hat{Q}_{\pi_{k-1}}(x, \cdot)\big\|_{\infty}^2\\
&+\sum_{k=1}^K 2\big\|Q_{\pi_k}(x, \cdot)- Q_{\pi_{k-1}}(x, \cdot)\big\|_{\infty}^2. 
    \end{split}
\end{equation}
The first two terms in Eq.~\eqref{eqn:Q_diff_dec} measure the estimation error. By Condition \ref{con:estimation_error}, we have,
\begin{equation}
\label{eq:ell_infty_err}
    \big\|\hat{Q}_{\pi_k}(x, \cdot)- Q_{\pi_k}(x, \cdot)\big\|^2_{\infty}\leq  \frac{ 2\tilde{C}^2 \log(1/\delta)}{\tau} + 2\varepsilon_0^2,
\end{equation}
with probability at least $1-\delta$ for each $k\in[K]$ and for problem-dependent constants $\tilde{C}$. Putting Eq. \eqref{eqn:Q_diff_dec}, Eq. \eqref{eq:ell_infty_err} and Lemma~\ref{lemma:Q_error} together, the following holds with probability $1-K\delta$,
\begin{eqnarray}\label{eqn:bound1}
&&\sum_{k=1}^K \big\|\hat{Q}_{\pi_k}(x, \cdot)- \hat{Q}_{\pi_{k-1}}(x, \cdot)\big\|_{\infty}^2\nonumber \\
&\leq& \frac{8\tilde{C}^2K\log(1/\delta)}{\tau} + 8K\varepsilon_0^2 +2\tmix^4\log_2^4(K) \sum_{k=1}^K\max_x\big\|\pi_{k}(\cdot|x) - \pi_{k-1}(\cdot|x)\big\|_1^2+\frac{4K}{K^6}.
\end{eqnarray}

\textbf{Step 3.} Finally, by our choice of the data-dependent learning rate $\eta_k(x)$, we are able to cancel out the positive term in Eq.~\eqref{eqn:R1_bound_conti} such that the regret is greatly sharpened. For notation simplicity, we denote $d_k(x) = \|\pi_k(\cdot|x)-\pi_{k-1}(\cdot|x)\|_1$. Putting Eq. \eqref{eqn:R1_bound_conti} and Eq. \eqref{eqn:bound1} together, with a union bound, we have 
\begin{equation}\label{eqn:bound2}
    \begin{split}
         \frac{R_{2T}}{\tau}\leq C_1\sum_{x\in\cX}\mu_{\pi^*}(x)\Big(\sqrt{R_{\max}}&\sqrt{ \tmix^4\log_2^4(K)\sum_{k=1}^K\max_xd_k^2(x)+\frac{\tilde{C}^2K\log(K\mu_{\pi^*}(x)\delta)^{-1}}{\tau} + K\varepsilon_0^2}\\
         &-\sum_{k=1}^K\frac{\eta_k(x)}{4}d_{k+1}^2(x) + 2(b+Q_{\max})\Big)
    \end{split}
\end{equation}
holds with probability at least $1-\delta$. Assuming $\eta_0(x) = \eta_1(x)$, we have 
\begin{equation*}
    \sum_{k=1}^K\frac{\eta_k(x)}{4}d_{k+1}^2(x)\geq \sum_{k=1}^K\frac{\eta_{k-1}(x)}{4}d_{k}^2(x).
\end{equation*}
Moreover, we denote $\mu^*_{\min} = \min_{x:\mu_{\pi^*}(x)>0} \mu_{\pi^*}(x)$ and 
\begin{equation*}
    \begin{split}
       &g_1=R_{\max}\tmix^4\log_2^4(K) \\
       &g_2 = \tilde{C}^2R_{\max}K\log(K\mu^*_{\min}\delta)^{-1}/\tau + K\varepsilon_0^2R_{\max}\\
       &g_3 = 2(b+Q_{\max}).
    \end{split}
\end{equation*}
Then we simplify Eq.~\eqref{eqn:bound2} as 
\begin{eqnarray*}
   \frac{R_{2T}}{\tau}&\leq& \sum_{x\in\cX}\mu_{\pi^*}(x)\Big(\sqrt{ g_1\sum_{k=1}^K\max_xd_k^2(x)+g_2}-\sum_{k=1}^K\frac{\eta_{k-1}(x)}{4}d_{k}^2(x) + g_3\Big)\\
   &=&\sqrt{ g_1\sum_{k=1}^K\max_xd_k^2(x)+g_2}-\sum_{x\in\cX}\mu_{\pi^*}(x)\sum_{k=1}^K\frac{\eta_{k-1}(x)}{4}d_{k}^2(x) + g_3.
\end{eqnarray*}
Let us denote $x_k^* = \argmax_xd_{k}^2(x)$. Noting that 
\begin{eqnarray}
  \sum_{x\in\cX}\mu_{\pi^*}(x)\sum_{k=1}^{K}\frac{\eta_{k-1}(x)}{4}d_{k}^2(x)\geq\sum_{k=1}^{K}\mu_{\pi^*}(x_k^*)\frac{\eta_{k-1}(x_k^*)}{4}d_{k}^2(x_k^*),
\end{eqnarray}
we have 
\begin{eqnarray*}
  \frac{R_{2T}}{\tau}&\leq& \sqrt{ g_1\sum_{k=1}^Kd_k^2(x_k^*)+g_2}-\sum_{k=1}^K\mu_{\pi^*}(x_k^*)\frac{\eta_{k-1}(x_k^*)}{4}d_{k}^2(x_k^*) + g_3\\
  &=& \sqrt{ g_1\sum_{k=1}^K\frac{\mu_{\pi^*}(x_k^*)}{\mu_{\pi^*}(x_k^*)}d_k^2(x_k^*)+g_2}-\sum_{k=1}^K\mu_{\pi^*}(x_k^*)\frac{\eta_{k-1}(x_k^*)}{4}d_{k}^2(x_k^*) + g_3\\
  &\leq& \sqrt{ \frac{4g_1}{\eta_{1}(x_k^*)\mu^*_{\min}}\sum_{k=1}^K\mu_{\pi^*}(x_k^*)\frac{\eta_{k-1}(x_k^*)}{4}d_k^2(x_k^*)+g_2}-\sum_{k=1}^K\mu_{\pi^*}(x_k^*)\frac{\eta_{k-1}(x_k^*)}{4}d_{k}^2(x_k^*) + g_3\\
  &=& 2\sqrt{ \frac{g_1}{\eta_{1}(x_k^*)\mu^*_{\min}}\Big(\sum_{k=1}^K\mu_{\pi^*}(x_k^*)\frac{\eta_{k-1}(x_k^*)}{4}d_k^2(x_k^*)+\frac{\mu^*_{\min}\eta_1g_2}{4g_1}\Big)}\\
 &&-\Big(\sum_{k=1}^K\mu_{\pi^*}(x_k^*)\frac{\eta_{k-1}(x_k^*)}{4}d_{k}^2(x_k^*) + \frac{\mu^*_{\min}\eta_{1}(x_k^*)g_2}{4g_1}\Big) +\frac{\mu^*_{\min}\eta_{1}(x_k^*)g_2}{4g_1}+ g_3,
\end{eqnarray*}
where the second inequality we use the fact that $\eta_k$ is monotone increasing. Letting 
\begin{eqnarray*}
   a =\frac{g_1}{\eta_{1}(x_k^*)\mu^*_{\min}}, b=\sum_{k=1}^{K}\mu_{\pi^*}(x_k^*)\frac{\eta_{k-1}(x_k^*)}{4}d_{k}^2(x_k^*)+ \frac{\mu^*_{\min}\eta_{1}(x_k^*)g_2}{4g_1},
\end{eqnarray*}
and using the fact that $2\sqrt{ab}-b \leq a$, we reach
\begin{equation}\label{eqn:bound_FTRL}
    \frac{R_{2T}}{\tau} \leq \frac{g_1}{\eta_{1}(x_k^*)\mu^*_{\min}} + \frac{\mu^*_{\min}\eta_{1}(x_k^*)g_2}{4g_1}+g_3.
\end{equation}

Plugging in back the definition of $g_1,g_2,g_2$, we have with probability at least $1-\delta$,
\begin{eqnarray}\label{eqn:bound_R_2T}
    R_{2T}\leq \frac{R_{\max}\tmix^4\log_2^4(K)\tau}{\eta_{1}(x_k^*)\mu^*_{\min}}+ \frac{\eta_{1}(x_k^*)(\tilde{C}^2K\log(K\mu^*_{\min}\delta)^{-1} + T\varepsilon_0^2)}{4\tmix^4\log_2^4(K)}+2(b+Q_{\max}).
\end{eqnarray}
By definition, $\eta_1(x_k^*) = \sqrt{2R_{\max}\|\hat{Q}_1(x_k^*, \cdot)\|_{\infty}}$. Since $\hat{Q}_1(x,a)\in[b,b+Q_{\max}]$ from Condition \ref{con:estimation_error}, we have $\eta_1(x_k^*)$ is lower and upper bounded by some constant. Based on this, we simplify the upper bound \eqref{eqn:bound_R_2T} as 
\begin{eqnarray*}
    R_{2T}\lesssim \frac{\tau\tmix^4\log_2^4(K)}{\mu_{\min}^*} +\tilde{C}^2K\log (K/\delta) + T\varepsilon_0^2,
\end{eqnarray*}
where $\lesssim$ hides constant factors.
This ends the proof.  \hfill $\blacksquare$\\

\subsection{Proof of Lemma \ref{lemma:Q_error}: relative $Q$-function error}\label{proof:relative_Q}
We first introduce a lemma that illustrates the true Q-value can be bounded by the mixing time.
\begin{lemma}[Lemma 3 in \cite{neu2010online}] \label{lemma:Q_bound}
For any policy $\pi$ and any state-action pair $(x, a)\in\cX\times \cA$, for any reward function $r \in [0, 1]$, we have 
\begin{equation}
    |Q_{\pi}(x, a)|\leq 2\tmix + 3.
\end{equation}
\end{lemma}

From the Bellman equation Eq.~\eqref{eqn:Bellman_eqn}, 
\begin{eqnarray}\label{eqn:Q_diff}
  Q_{\pi_k}(x, a) - Q_{\pi_{k-1}}(x, a)
  = \sum_{x'}\cP(x'|x,a)\Big(V_{\pi_k}(x')-V_{\pi_{k-1}}(x')\Big) + \lambda_{\pi_{k-1}}-\lambda_{\pi_k}.
\end{eqnarray}
We first bound $\lambda_{\pi_{k-1}}-\lambda_{\pi_k}$. By Lemma \ref{lemma_per_diff} (performance difference lemma), 
\begin{equation*}
    \lambda_{\pi_{k-1}}-\lambda_{\pi_{k}} = \sum_{x}\mu_{\pi_{k-1}}(x)\Big(\sum_{a}(\pi_{k-1}(a|x)-\pi_{k}(a|x))\Big)Q_{\pi_{k}}(x,a).
\end{equation*}
By Lemma \ref{lemma:Q_bound}, it implies
\begin{equation}\label{eqn:bound_lambda}
    \lambda_{\pi_{k-1}}-\lambda_{\pi_{k}}\leq (2\tmix+3)\max_{x}\big\|\pi_{k-1}(\cdot|x)-\pi_k(\cdot|x)\big\|_1.
\end{equation}

Next we bound $V_{\pi_k}(x)-V_{\pi_{k-1}}(x)$. In an ergodic MDP, the expected average reward $\lambda_{\pi}$ can be written as $\lambda_{\pi} = \mu_{\pi}^{\top}r_{\pi}$, where $r_{\pi}(x) = \sum_a\pi(a|x)r(x,a)$. Let $\be_x$ be an indicator vector for state $x$. For all $\pi$,
\begin{eqnarray}\label{eqn:V_dec}
  V_{\pi}(x) &=& \sum_{t=0}^{\infty}\Big(\be_x^{\top}(\cP^{\pi})^t-\mu_{\pi}^{\top}\Big)r_\pi \nonumber\\
  &=& \sum_{t=0}^{N-1}\Big(\be_x^{\top}(\cP^{\pi})^t-\mu_{\pi}^{\top}\Big)r_{\pi} + \sum_{t=N}^{\infty}\Big(\be_x^{\top}(\cP^{\pi})^t-\mu_{\pi}^{\top}\Big)r_{\pi},
\end{eqnarray}

Corollary 13.2 of \citet{wei2019modelfree} shows that for an ergodic MDP with mixing time $\tmix$ and $N = \lceil 4\tmix\log_2(K) \rceil$, for all $\pi$,
\begin{eqnarray*}
  \sum_{t=N}^{\infty} \Big\|\be_x^{\top}(\cP^{\pi})^t-\mu_{\pi}\Big\|_1\leq \sum_{t=N}^{\infty}2^{1-\frac{t}{\tmix}}=\frac{2^{1-\frac{N}{\tmix}}}{1-2^{-\frac{1}{\tmix}}}\leq \frac{2\tmix}{\ln 2}2^{1-\frac{N}{\tmix}}=\frac{2\tmix}{\ln 2}\frac{2}{K^4}\leq \frac{1}{K^3}.
\end{eqnarray*}
Thus, the second term in Eq.~\eqref{eqn:V_dec} can be bounded by
\begin{equation*}
    \Big|\sum_{t=N}^{\infty}\Big(\be_x^{\top}(\cP^{\pi})^t-\mu_{\pi}^{\top}\Big)r_{\pi}\Big|\leq \sum_{t=N}^{\infty} \Big\|\be_x^{\top}(\cP^{\pi})^t-\mu_{\pi}\Big\|_1\leq \frac{1}{K^3}.
\end{equation*}

The following steps are similar to the proof of Lemma 7 in \citet{wei2019modelfree}. For the sake of completeness, we present a full proof here. The difference between $V_{\pi_k}(x)$ and $V_{\pi_{k-1}}(x)$ can be bounded by
\begin{eqnarray}\label{eqn:V_diff}
  &&\Big|V_{\pi_k}(x)-V_{\pi_{k-1}}(x)\Big|\nonumber\\
  &=&\Big|\sum_{t=0}^{N-1}\be_x^{\top}\Big((\cP^{\pi_k})^t-(\cP^{\pi_{k-1}})^t\Big)r_{\pi_k}+\sum_{t=0}^{N-1}\be_x^{\top}(\cP^{\pi_k})^t(r_{\pi_k}-r_{\pi_{k-1}})-N\lambda_{\pi_k}+N\lambda_{\pi_{k-1}}\Big|+  \frac{2}{K^3} \nonumber\\
  &\leq&\sum_{t=0}^{N-1}\Big\|\Big((\cP^{\pi_k})^t-(\cP^{\pi_{k-1}})^t\Big)r_{\pi_k}\Big\|_{\infty} + \sum_{t=0}^{N-1}\|r_{\pi_k}-r_{\pi_{k-1}}\|_{\infty}+N|\lambda_{\pi_k}-\lambda_{\pi_{k-1}}|+  \frac{2}{K^3}.
\end{eqnarray}
Next, we will derive a recursive form for the first term as follows:
\begin{eqnarray*}
  &&\Big\|\Big((\cP^{\pi_k})^t-(\cP^{\pi_{k-1}})^t\Big)r_{\pi_k}\Big\|_{\infty}\\
  &=&\Big\|\Big(\cP^{\pi_k}(\cP^{\pi_k})^{t-1}-\cP^{\pi_k}(\cP^{\pi_{k-1}})^{t-1}+ \cP^{\pi_k}(\cP^{\pi_{k-1}})^{t-1}-\cP^{\pi_{k-1}}(\cP^{\pi_{k-1}})^{t-1}\Big)r_{\pi_k}\Big\|_{\infty}\\
  &\leq& \Big\|\cP^{\pi_k}\Big((\cP^{\pi_k})^{t-1}-(\cP^{\pi_{k-1}})^{t-1}\Big)r_{\pi_k}\Big\|_{\infty}+ \Big\|(\cP^{\pi_k}-\cP^{\pi_{k-1}})(\cP^{\pi_{k-1}})^{t-1}r_{\pi_k}\Big\|_{\infty}\\
  &\leq&\Big\|\Big((\cP^{\pi_k})^{t-1}-(\cP^{\pi_{k-1}})^{t-1}\Big)r_{\pi_k}\Big\|_{\infty}+\max_x\Big\|\be_x^{\top}(\cP^{\pi_k}-\cP^{\pi_{k-1}})(\cP^{\pi_{k-1}})^{t-1}\Big\|_1\\
  &\leq&\Big\|\Big((\cP^{\pi_k})^{t-1}-(\cP^{\pi_{k-1}})^{t-1}\Big)r_{\pi_k}\Big\|_{\infty}+\max_x\Big\|\be_x^{\top}(\cP^{\pi_k}-\cP^{\pi_{k-1}})\Big\|_1\\
  &\leq&\Big\|\Big((\cP^{\pi_k})^{t-1}-(\cP^{\pi_{k-1}})^{t-1}\Big)r_{\pi_k}\Big\|_{\infty}+\max_x\Big(\sum_{x'}\Big|\sum_a\Big(\pi_k(a|x)-\pi_{k-1}(a|x)\Big)\cP(x'|x,a)\Big|\Big)\\
  &\leq&\Big\|\Big((\cP^{\pi_k})^{t-1}-(\cP^{\pi_{k-1}})^{t-1}\Big)r_{\pi_k}\Big\|_{\infty}+\max_x\Big\|\pi_k(a|x)-\pi_{k-1}(a|x)\Big\|_1.
\end{eqnarray*} 
By induction, it holds that 
\begin{equation*}
    \Big\|\Big((\cP^{\pi_k})^t-(\cP^{\pi_{k-1}})^t\Big)r_{\pi_k}\Big\|_{\infty}\leq t\max_x\Big\|\pi_k(a|x)-\pi_{k-1}(a|x)\Big\|_1.
\end{equation*}
Thus,
\begin{equation}\label{eqn:bound_I1}
    \sum_{t=0}^{N-1}\Big\|\Big((\cP^{\pi_k})^t-(\cP^{\pi_{k-1}})^t\Big)r_{\pi_k}\Big\|_{\infty} \leq N^2 \max_x\Big\|\pi_k(a|x)-\pi_{k-1}(a|x)\Big\|_1.
\end{equation}
In addition,
\begin{equation}\label{eqn:bound_I2}
    \sum_{t=0}^{N-1}\|r_{\pi_k}-r_{\pi_{k-1}}\|_{\infty}\leq N\max_x\Big\|\pi_k(a|x)-\pi_{k-1}(a|x)\Big\|_1.
\end{equation}
Plugging Eq.~\eqref{eqn:bound_lambda}, Eq.~\eqref{eqn:bound_I1} and Eq.~\eqref{eqn:bound_I2} into Eq.~\eqref{eqn:V_diff} yields
\begin{equation}
    \Big|V_{\pi_k}(x)-V_{\pi_{k-1}}(x)\Big| \leq \Big(N^2+N+(2\tmix+3)N\Big)\max_x\Big\|\pi_k(a|x)-\pi_{k-1}(a|x)\Big\|_1+  \frac{2}{K^3},
\end{equation}
where $N = \lceil 4\tmix\log_2(K) \rceil$. Together with Eq.~\eqref{eqn:Q_diff}, we reach the result. \hfill $\blacksquare$\\

\section{Linear value function approximation}
\label{app:ell_infty}

In this section, we show that with linear value function approximation and under similar assumptions as in \citet{lazic2019politex}, the estimation error in each state can be bounded in the $\ell_{\infty}$ norm. Note that we consider an unrealizable case such that Q-function could be approximated linear represented up to an irreducible approximation error $\varepsilon_0$. This is in contrast of many existing works \citep{yang2019sample, yang2019reinforcement, jin2019provably} who consider realizable cases. 

Suppose $\phi:\cX\times \cA\to \mathbb R^d$ is a feature map chosen by the user. Consider $\widehat Q_{\pi_k} (x, a) = \phi(x, a)^\top \widehat w_k$ be the linear value function estimate where $\hat{w}_k$ is the estimated weight vector. Let $\Psi$ be a $|\cX||\cA|\times d$ feature matrix whose rows correspond to state-action feature vectors. We make the regularity assumption on $\Psi$ and assume that for all policies $\pi$, the following feature excitation condition holds.
\begin{assumption}[Linearly independent features]\label{assum:linear_independent}
The columns of the matrix $[\Psi, \bm{1}]$ are linearly independent. 
\end{assumption}
\begin{assumption}[Uniformly excited features, Assumption A4 in \citet{lazic2019politex}]
\label{assumption:exploration-all}
There exists a positive real $\sigma$ such that for any policy $\pi$, $\lambda_{\min}(\Psi^\top {\rm diag}(\mu_\pi\otimes \pi) \Psi) \ge \sigma$.
\end{assumption}
Furthermore, we assume that the following error bound holds.
\begin{assumption}[Estimation error in $\mu_\pi\otimes \pi$-norm]
\label{assumption:weighted-error}
For all $k \in [K]$, with probability at least $1-\delta$, the value error is bounded in the $\mu_\pi\otimes \pi$-norm.
\begin{equation*}
\Big\| \widehat{Q}_{\pi_k} - Q_{\pi_k} \Big\|_{\mu_\pi\otimes \pi}  \leq C_2 \sqrt{\frac{\log(1/\delta)}{\tau}} + \varepsilon_0 \,,
\end{equation*}
where $C_2$ is a problem-dependent constant and $\varepsilon_0$ is the irreducible approximation error. 
\end{assumption}
The above error Assumption \ref{assumption:weighted-error} can be satisfied, for example, by the LSPE algorithm of \citet{bertsekas1996temporal}, as shown in Theorem 5.1 in \citet{lazic2019politex}. The same authors show that Assumptions \ref{assum:linear_independent}, \ref{assumption:exploration-all} and \ref{assumption:weighted-error} also suffice to bound the error in $\mu^* \otimes \pi_k$ and $\mu^*\otimes \pi^*$-norms, as required by our Lemma~\ref{lemma:estimation_error}. Here we additionally prove that under same assumptions, the error in each state is bounded in the $\ell_{\infty}$-norm, as required by Lemma~\ref{lemma:1}.

\begin{lemma}[Estimation error in $\ell_{\infty}$-norm]\label{lemma:1}
Under Assumptions \ref{assumption:exploration-all} and \ref{assumption:weighted-error}, we have the following holds with probability at least $1-\delta$,
\begin{equation*}
    \big\|\widehat{Q}_{\pi_k}(x, \cdot)- Q_{\pi_k}(x, \cdot)\big\|_{\infty}\leq  C_{\psi} \big( C_2 \sqrt{\sigma \frac{\log(1/\delta)}{\tau}} + \varepsilon_0 \big),
\end{equation*}
where $C_\Psi = \max_{x, a}\|\psi(x, a)\|_2$. 
\end{lemma}

\textbf{Proof.} Note that under Assumption~\ref{assumption:exploration-all}, $\norm{\Psi (\widehat w_k - w_k)}_{\mu_{\pi}\otimes \pi}^2 \geq \sigma \norm{\widehat w_k - w_k}^2_2 $.  We have the following:
\begin{align*}
\| \widehat Q_{\pi_k}(x, \cdot) - Q_{\pi_k}(x, \cdot) \|_{\infty} & = \max_a | \phi(x, a)^\top (\widehat w_k - w_k) | \\
& \leq C_{\Psi} \norm{ \widehat w_k - w_k}_2 \\
& \leq C_{\Psi}  \norm{\Psi (\widehat w_k - w_k)}_{\mu_{\pi}\otimes \pi}/\sqrt{\sigma} \\
& = C_{\Psi} \| \widehat{Q}_{\pi_k} - Q_{\pi_k} \|_{\mu_{\pi}\otimes \pi}/\sqrt{\sigma} \, \\
& \leq C_{\Psi}C_2 \sqrt{\log (1/\delta)/ (\sigma\tau)} + C_{\Psi}/\sqrt{\sigma} \varepsilon_0 \,.
\end{align*}
\hfill $\blacksquare$\\

\section{Supporting lemmas}\label{sec:supporting}

\begin{lemma}[Performance difference lemma]\label{lemma_per_diff}
Consider an MDP specified by the transition probability kernel $\cP$ and reward function $r$.
For any policy $\pi,\hat{\pi}$, it holds that
\begin{equation*}
    \lambda_{\pi} -\lambda_{\hat{\pi}} = \sum_{x, a}\mu_{\pi}(x)(\pi(a|x)-\hat{\pi}(a|x))Q_{\hat{\pi}}(x, a),
\end{equation*}
where $\mu_{\pi}(x)$ is the stationary distribution of a policy $\pi$.
\paragraph{Proof.} Based on average reward Bellman equation, we have
\begin{eqnarray*}
  \sum_{x,a}\mu_{\pi}(x)\pi(a|x)Q_{\hat{\pi}}(x,a)&=& \sum_{x,a}\mu_{\pi}(x)\pi(a|x)\Big[r(x,a)-\lambda_{\hat{\pi}}+\sum_{x'}\cP(x'|x,a)V_{\hat{\pi}}(x')\Big]\\
  &=&\lambda_{\pi}-\lambda_{\hat{\pi}} + \sum_{x}\mu_{\pi}(x)V_{\hat{\pi}}(x),
\end{eqnarray*}
where the second equation is due to $\sum_{x,a}\mu_{\pi}(x)\pi(a|x)\cP(x'|x,a) = \mu_{\pi}(x')$. Therefore,
\begin{eqnarray*}
  \lambda_{\pi}-\lambda_{\hat{\pi}} &=& \sum_{x,a}\mu_{\pi}(x)\pi(a|x)Q_{\hat{\pi}}(x,a) - \sum_{x}\mu_{\pi}(x)V_{\hat{\pi}}(x)\\
  &=&\sum_{x,a} \mu_{\pi}(x)\Big(\pi(a|x)Q_{\hat{\pi}}(x,a)-\hat{\pi}(a|x)Q_{\hat{\pi}}(x,a)\Big).
\end{eqnarray*}
This ends the proof. \hfill $\blacksquare$\\

\begin{lemma}[Lemma 4 in \cite{mcmahan2017survey}]\label{lemma:au}
For any non-negative real numbers $a_1, \ldots, a_T$, the following holds
\begin{equation*}
    \sum_{t=1}^T\frac{a_t}{\sqrt{\sum_{s=1}^ta_s}}\leq 2\sqrt{\sum_{t=1}^Ta_t}.
\end{equation*}
\end{lemma}
\end{lemma}

%% file: ftrl.tex
In Section \ref{sec:proof_main}, we present the detailed proofs of main results. In Section \ref{app:ell_infty}, the linear value function approximation is considered. In Section \ref{sec:supporting}, some supporting lemmas are included. 

\section{Proofs of main results}\label{sec:proof_main}
\subsection{Proof of Theorem \ref{thm:main}: main result}

We combine the decomposition \eqref{eqn:regret_dec}, \eqref{eqn:per_diff} and \eqref{eqn:decom_I1} together and utilize the results in Lemmas \ref{lemma:bound_VW}, \ref{lemma:estimation_error} and \ref{thm:R_2T}. Then we have
\begin{equation*}
    \begin{split}
R_T\lesssim& \underbrace{K\tmix + 4K\tmix\sqrt{2\tau\log (T/\delta)}}_{\text{Lemma} \ \ref{lemma:bound_VW}} + \underbrace{\tilde{C}T\sqrt{\frac{\log(1/\delta)}{\tau}} + T\varepsilon_0 }_{\text{Lemma} \ \ref{lemma:estimation_error}}\\
&+ \underbrace{\frac{\tau t_{\max}^4\log_2^4(K)}{\mu_{\min}^*}+T\Big(\frac{\tilde{C}^2\log(1/\delta)}{\tau}+\varepsilon_0^2\Big)}_{\text{Lemma} \ \ref{thm:R_2T}}.
    \end{split}
\end{equation*}
We choose $\delta = 1/T$ and ignore any universal constant and logarithmic factor in the following. Since $K = T/\tau$, it holds that 
\begin{equation*}
    \begin{split}
        R_T &\overset{\log}{\lesssim} \tmix K\tau^{1/2}+ \tilde{C}T\tau^{-1/2}+\frac{\tmix^4\tau}{\mu_{\min}^*}+ T(\varepsilon_0^2+\varepsilon_0)\\
        &\overset{\log}{\lesssim}\tmix\tilde{C}\tau^{-1/2}T+\frac{\tmix^4\tau}{\mu_{\min}^*} + T(\varepsilon_0^2+\varepsilon_0),
    \end{split}
\end{equation*}
with probability at least $1-1/T$. With a little abuse of notations, we re-define $\varepsilon_0 = \varepsilon_0^2+\varepsilon_0$. By optimizing $\tau$ such that the first two term above is equal, i.e., $ \tmix\tilde{C}\tau^{-1/2}T=\tmix^4\tau/\mu_{\min}^*$,
we choose $\tau = (\tilde{C}\mu_{\min}^*/\tmix^3)^{2/3}T^{2/3}$. Overall, we  reach the final regret bound,
$$
R_T = \tilde{\cO}\left(\tmix^2\rho^{1/3}\tilde{C}^{2/3}T^{2/3} + T\varepsilon_0\right),
$$ 
where $\rho = \max_{\pi}\max_{x:\mu_{\pi}(x)\neq 0} (1/\mu_{\pi}(x))$. This ends the proof.

\hfill $\blacksquare$\\

\subsection{Proof of Lemma \ref{lemma:FTRL}: adaptive optimistic FTRL (AO-FTRL)}
\label{proof:FTRL}
Lemma~\ref{lemma:FTRL} states that the cumulative regret for AO-FTRL is upper-bounded by
\begin{equation*}
\begin{split}
     R_T&\leq \Big(\frac{8}{\eta}+\eta\cR(f^*)\Big) \sqrt{\sum_{t=2}^T\|q_t-M_t\|^2_{*}}
     -\sum_{t=1}^{T}\frac{\eta_{t}}{4}\|f_t-f_{t+1}\|^2 + g,
\end{split}
\end{equation*}
where $g = \langle M_{T+1}, f^*-f_{T+1}\rangle + \|q_1\|_{*}^2/\eta_1$.

First, at each round $t$, AO-FTRL has the form of
\begin{equation*}
    \begin{split}
         f_{t+1} &= \argmin_{f\in\cF}\langle f, \sum_{s=1}^t q_s + M_{t+1}\rangle + \eta_t \cR(f)\\
    &=\argmin_{f\in\cF} \langle f, \sum_{s=1}^tq_s\rangle + \sum_{s=1}^t\langle M_{s+1}-M_s, f\rangle + \eta_t\cR(f),
    \end{split}
\end{equation*}
where $\eta_1\leq \cdots\leq \eta_t$ are data-dependent learning rates. For simplicity, we assume $\eta_0 = 0$. For $s=1, \ldots, t$, we define
\begin{equation}\label{def:q}
    h_s(f) = \langle M_{s+1} - M_s, f\rangle + (\eta_s-\eta_{s-1})\cR(f).
\end{equation}
We define $h_0(f)=0$ for all $f\in\cF$ and $h_{1:t}(f)=\sum_{s=1}^t h_s(f) = \langle M_{t+1}, f \rangle + \eta_t R(f)$. Since $\cR(f)$ is 1-strongly convex with respect to norm $\|\cdot\|$, $h_s(f)$ is  $(\eta_s-\eta_{s-1})$-strongly-convex with respect to $\|\cdot\|$. Then we could rewrite the AO-FTRL update as
\begin{equation*}
    f_{t+1} = \argmin_{f\in\cF} \langle f, \sum_{s=1}^t q_s\rangle + \sum_{s=1}^t h_s(f).
\end{equation*}

Second, let us define the forward linear regret $R_T^+$ as:
\begin{equation*}
    R_T^+ = \sum_{t=1}^T\big\langle q_t, f_{t+1}-f^*\big\rangle.
\end{equation*}
One could interpret $R_T^+$ as a cheating regret since it uses prediction
$f_{t+1}$ at round $t$. We decompose the cumulative regret based on the forward linear regret as follows,
\begin{equation}\label{eqn:regret_decom}
    R_T = \sum_{t=1}^T\langle q_t,f_t\rangle - \sum_{t=1}^T\langle q_t,f^*\rangle= R_T^+ + \sum_{t=1}^T\langle q_t, f_t-f_{t+1} \rangle.
\end{equation}
The second term in the right side captures the regret by the algorithm’s inability to accurately predict the future. We define the Bregman divergence between two
vectors induced by a differentiable function $R$ as follows:
\begin{equation*}
    \cD_{R}(w,u) = R(w)-\Big(R(u) + \langle \nabla R(u), w-u\rangle\Big).
\end{equation*}
Next theorem is used to bound the forward regret.
\begin{theorem}[Theorem 3 in \cite{joulani2017modular}]\label{thm:forward_regret}
For any $f^*\in\cF$ and any sequence of linear losses, the forward regret satisfies the following inequality:
\begin{equation*}
    R_T^+ \leq \sum_{t=1}^T \Big(h_t(f^*)-h_t(f_{t+1})\Big)-\sum_{t=1}^T\cD_{h_{1:t}}(f_{t+1},f_t).
\end{equation*}
\end{theorem}
Recall that $h_{1:t}(f)$ is $\eta_t$-strongly convex. From the definitions of strong convexity and Bregman divergence, we have 
\begin{equation}\label{eqn:breg_diverg}
    \sum_{t=1}^T\cD_{h_{1:t}}(f_{t+1},f_t)\geq \sum_{t=1}^T\frac{\eta_t}{2}\|f_{t+1}-f_t\|^2.
\end{equation}
Applying Theorem \ref{thm:forward_regret} and Eq.~\eqref{eqn:breg_diverg}, we have
\begin{eqnarray}\label{eqn:forward_decom}
   R_T^+ \leq \sum_{t=1}^T\Big(h_t(f^*)-h_t(f_{t+1})\Big) -\sum_{t=1}^T\frac{\eta_t}{2}\|f_{t+1}-f_{t}\|^2.
\end{eqnarray}
To bound the first term in Eq.~\eqref{eqn:forward_decom}, we expand it by the definition of Eq.~\eqref{def:q}, 
\begin{eqnarray}\label{eqn:expand}
   && \sum_{t=1}^T\Big(h_t(f^*)-h_t(f_{t+1})\Big) \nonumber\\
    &=& \sum_{t=1}^T\langle M_{t+1}-M_t, f^*-f_{t+1}\rangle + \sum_{t=1}^T(\eta_t-\eta_{t-1})(\cR(f^*)-\cR(f_{t+1}))\nonumber\\
    &\leq& \sum_{t=1}^T\langle M_{t+1}-M_t, f^*\rangle  -\sum_{t=1}^T\langle M_{t+1}-M_t, f_{t+1}\rangle +  \eta_T\cR(f^*)\nonumber\\
    &=& \langle M_{T+1}, f^*\rangle  -\sum_{t=1}^T\langle M_{t+1}-M_t, f_{t+1}\rangle +  \eta_T\cR(f^*),
\end{eqnarray}
where the first inequality is due to the fact that $\eta_t$ is non-decreasing and $\eta_0=0$. We decompose the second term in Eq.~\eqref{eqn:expand} as follows,
\begin{eqnarray}\label{eqn:expand2}
    \sum_{t=1}^T\langle M_{t+1}-M_t, f_{t+1}\rangle &=& \sum_{t=2}^{T+1}\langle M_t, f_t\rangle - \sum_{t=1}^T\langle M_t, f_{t+1}\rangle\nonumber\\
    &=&\sum_{t=1}^T\langle M_t,f_t\rangle-\sum_{t=1}^T\langle M_t, f_{t+1}\rangle +\langle M_{T+1}, f_{T+1}\rangle,
\end{eqnarray}
since $M_1=0$. Combining Eq.~\eqref{eqn:expand} and Eq.~\eqref{eqn:expand2} together, 
\begin{equation}\label{eqn:h_regu}
     \sum_{t=1}^T\Big(h_t(f^*)-h_t(f_{t+1})\Big) = -\sum_{t=1}^T\langle M_t, f_t-f_{t+1}\rangle + \langle M_{T+1}, f^*-f_{T+1}\rangle +  \eta_T\cR(f^*).
\end{equation}

Putting Eq.~\eqref{eqn:regret_decom}, Eq.~\eqref{eqn:forward_decom} and Eq.~\eqref{eqn:h_regu} together, we reach
\begin{eqnarray}\label{eqn:R_bound}
    R_T\leq \sum_{t=1}^T\langle q_t-M_t, f_t-f_{t+1}\rangle - \sum_{t=1}^T\frac{\eta_t}{2}\|f_t-f_{t+1}\|^2+ \langle M_{T+1}, f^*-f_{T+1}\rangle  + \eta_T\cR(f^*).
\end{eqnarray}
To bound the first term in Eq.~\eqref{eqn:R_bound}, we first use H\"older's inequality such that
\begin{eqnarray*}
    \langle q_t-M_t, f_t-f_{t+1}\rangle 
    &=&\frac{2}{\eta_t}(q_t-M_t)^{\top}\frac{\eta_t}{2}(f_t-f_{t+1})\\
    &\leq& \|\frac{2}{\eta_t}(q_t - M_t)\|_* \|\frac{\eta_t}{2}(f_t - f_{t+1})\|\\
    &\leq&\frac{1}{\eta_t}\|q_t-M_t\|_{*}^2+\frac{\eta_t}{4}\|f_t-f_{t+1}\|^2,
\end{eqnarray*}
where the last inequality is due to $2ab \leq a^2 + b^2$. Thus we have
    \begin{eqnarray*}
     R_T&\leq& \sum_{t=1}^{T}\frac{1}{\eta_t}\|q_t-M_t\|_{*}^2+\sum_{t=1}^{T}\frac{\eta_t}{4}\|f_t-f_{t+1}\|^2 - \sum_{t=1}^T\frac{\eta_t}{2}\|f_t-f_{t+1}\|^2\\
    &&+ \langle M_{T+1}, f^*-f_{T+1}\rangle  + \eta_T\cR(f^*)\\
    &=&\sum_{t=1}^{T}\frac{1}{\eta_t}\|q_t-M_t\|_{*}^2-\sum_{t=1}^{T}\frac{\eta_t}{4}\|f_t-f_{t+1}\|^2 + \langle M_{T+1}, f^*-f_{T+1}\rangle  + \eta_T\cR(f^*).
\end{eqnarray*}
By choosing $\eta_t = \eta\sqrt{\sum_{s=1}^t\|q_s-M_s\|^2_{*}}$ for some absolute constant $\eta$, we have
\begin{equation}\label{eqn:decomp1}
    \begin{split}
         R_T\leq \sum_{t=1}^T\frac{\|q_t-M_t\|_{*}^2}{\eta\sqrt{\sum_{s=1}^t\|q_t-M_t\|_{*}^2}} &+\eta\sqrt{\sum_{t=1}^T\|q_t-M_t\|_{*}^2} \cR(f^*)\\
         &-\sum_{t=1}^{T}\frac{\eta_{t}}{4}\|f_t-f_{t+1}\|^2 + \langle M_{T+1}, f^*-f_{T+1}\rangle.
    \end{split}
\end{equation}
Lemma 4 in \citet{mcmahan2017survey} states that 
for any non-negative real numbers $a_1, \ldots, a_T$, 
$$
\sum_{t=1}^T\frac{a_t}{\sqrt{\sum_{s=1}^ta_s}}\leq 2\sqrt{\sum_{t=1}^Ta_t}.
$$
Applying this inequality to the first term in Eq.~\eqref{eqn:decomp1}  with $a_t=\|q_t-M_t\|_{*}^2$, we have
\begin{equation*}
    R_T\leq \Big(\frac{2}{\eta}+\eta\cR(f^*)\Big) \sqrt{\sum_{t=1}^T\|q_t-M_t\|^2_{*}} -\sum_{t=1}^{T}\frac{\eta_{t}}{4}\|f_t-f_{t+1}\|^2 + \langle M_{T+1}, f^*-f_{T+1}\rangle.
\end{equation*}
Letting $\eta = \sqrt{2/\cR(f^*)}$ and $R_{\max} = \max_{f}\cR(f)$, this concludes the proof. 
\hfill $\blacksquare$\\